\documentclass[conference]{IEEEtran}
\ifCLASSOPTIONcompsoc
  \usepackage[nocompress]{cite}
\else
  \usepackage{cite}
\fi
\ifCLASSINFOpdf
  
\else
  
\fi
\usepackage{amsmath}
\usepackage{graphicx}
\newcommand{\norm}[1]{\left\lVert#1\right\rVert}

\usepackage[usenames, dvipsnames]{color}
\usepackage[colorinlistoftodos]{todonotes}
\usepackage{subfiles}
\usepackage{blindtext}
\usepackage{algorithm2e}
\usepackage{hyperref}
\usepackage{color}
\usepackage{amsfonts}
\usepackage{textcomp}
\def\BibTeX{{\rm B\kern-.05em{\sc i\kern-.025em b}\kern-.08em
    T\kern-.1667em\lower.7ex\hbox{E}\kern-.125emX}}
\IEEEoverridecommandlockouts

\begin{document}

\title{Modeling and Experimental Validation of the Mechanics of a Wheeled Non-Holonomic Robot Capable of Enabling Homeostasis\\
\thanks{This work is supported by the Georgia Institute of Technology and the National Science Foundation}
}

\author{
\IEEEauthorblockN{Jeremy Epps}
\IEEEauthorblockA{School of Aerospace Engineering\\
Georgia Institute of Technology\\
Atlanta, Georgia 30332--0250\\
jeremy.epps@gatech.edu
}
\and
\IEEEauthorblockN{Dr. Eric Feron}
\IEEEauthorblockA{School of Aerospace Engineering\\
Georgia Institute of Technology\\
Atlanta, Georgia 30332--0250\\
eric.feron@aerospace.gatech.edu
}

\and
\IEEEauthorblockN{Mark Mote}
\IEEEauthorblockA{School of Aerospace Engineering\\
Georgia Institute of Technology\\
Atlanta, Georgia 30332--0250\\
mmote3@gatech.edu
}
}

\maketitle

\begin{abstract}
The field of bio-inspired robotics seeks to create mechanical systems that mimic the designs and concepts used by biological systems.  One of the more challenging biological concepts to imitate in mechanical systems is the ability to create an internal environment that can foster homeostasis through the use of a membrane similar to skin. A robot with this ability would be able to regulate internal parameters, repair itself using the internal sub-environment, and defend its internal parts from the surrounding environment. This paper presents the internal structure of a non-holonomic wheeled system that enables homeostasis via a fully connected interior, protected from the outside environment by a flexible membrane. The three objectives of this paper are to: 1) Explore the idea of nature creating higher-order life forms with wheeled limbs given the correct intermediate steps. 2) Characterize a robot that uses a homeostasis enabling wheel. 3) Determine the feasibility of using a homeostasis enabling wheel as a mode of locomotion.   
\end{abstract}


%
\IEEEpeerreviewmaketitle

\section{Introduction}

Homeostasis, in a biological context, is the ability of a biological system to coordinate a physiological response that maintains specific internal parameters at a steady state despite the existence of external disturbances \cite{cannon_1929:1}. Biological systems, such as the human body, rely on a membrane layer (skin, nails, and hair) to enable homeostasis.  The membrane layer creates a distinguishable, fully connected interior and protects all of the system's internals from the outside environment.  This membrane layer also aids biological systems in the process of self-repair through use of the internal sub-region\cite{starr_2005}.

In the field of robotics, homeostasis can have a similar definition as to when mentioned in a biological context. In robotics, depending on the parameters of interest, this definition can be satisfied by a system that does not create a sub-environment. The presented research focuses on homeostasis in the context of maintaining internal parameters such as temperature, pressure, and relative lubrication. Therefore, a two-dimensional membrane is needed so that the system creates two non-intersecting sub-environments allowing homeostasis to be feasible. Moreover, the mechanics of the system must be constrained such that the membrane remains intact while the system performs its tasks.  

This paper presents a novel non-holonomic wheeled robotic design that has the necessary structure and mechanics to enable homeostasis. This design allows for several advantages that the traditional wheel and axle designs do not have. Similar to a biological system, the presented robotic design creates a sub-environment necessary for the robot to repair itself and regulate internal parameters via its fully connected interior. These capabilities can be attributed to the continuous membrane that encloses the systems internal parts. The fully connected membrane would remain intact throughout the use of the robot due to the robots unique mechanics.  The design also allows for a unique range of motion in comparison to the traditional wheel and axle design used by cars and robots alike. 

The argument of whether or not nature has created the wheel is one of differing views\cite{Duke}\cite{labarbera_1983}\cite{biopolyverse_2015}. One of the more prevalent explanations as to why nature has not "invented" the wheel is that the Earth's natural environment is not conducive to wheeled locomotion. Evidence has been presented that wheeled locomotion was voluntarily abandoned in the desert's of Africa due to its ineffectiveness in the desert terrain\cite{diamond_1983}\cite{bulliet_1975}.  However, this view is not supported by the fact that wheeled robotics are used to explore the desert-like terrains of the Moon and Mars by organizations such as the Jet Propulsion Laboratory and the National Aeronautics and Space Administration\cite{young_2007}\cite{squyres_2006}. The inability of humankind's continuously rotating wheel to create a fully connected interior with the system it propels has also been listed as a primary reason why nature has not "invented" a wheeled limb\cite{wolchover_2012}\cite{diamond_1983}. The robot presented in this paper aims to negate this idea. 

There are no engineered systems with rotating elements that are designed in a manner that creates a fully connected internal area, as previously discussed. However, there are some wheels, found in the automotive industry, that are designed to regulate their own internal parameters. Several technologies \cite{benedict} \cite{wadmare_pandure_2017} \cite{Rubber}\cite{Tyre} have been patented that address the issue of regulating air pressure inside of a tire as well as the self-repair of a punctured tire. Thus the principle of homeostasis is not foreign in the field of mechanical systems and robotics. Previous research involving social robots focused on maintaining collective energy homeostasis between a team of robots \cite{Zhou:2013:EPS:2568493.2569289}. Research on homeostasis in electrical systems, inspired by biological immune systems, has also been a topic of interest\cite{10.1007/978-3-540-73922-7_19}. In \cite{10.1007/978-3-540-73922-7_19}, Owens discusses the architecture of a control system designed to maintain homeostasis of arbitrary parameters. The research also highlights the necessity of sensors, homeostatic variables, actuators, etc. in regard to creating a system that maintains homeostasis. None of the mentioned research involves maintaining homeostasis in a mechanical system, as proposed in this paper. The presented research builds on prior research conducted in the Georgia Institute of Technology\textquotesingle s Decision and Control Laboratory\cite{DBLP:journals/corr/AfmanMF17}\cite{epps_feron_2019}. The main contributions of this paper are summarized as follows:

\begin{itemize}
\item Description of the robot and characterization of the homeostasis enabling wheel  
\item Defining the general coordinates and kinematics of the robot 
\item Experimental validation of the feasibility of a robot using a homeostasis enabling wheel
\item An evolutionary consideration of the creation of a homeostasis enabling wheel
\end{itemize}


\section{Description of the Robot, its Motion and its Working Principles}

The robot created through this research resembles a traditional tricycle in the sense that it makes use of three wheels, two rear wheels and one front wheel. The two rear wheels provide stability for the robot and are not powered while the front wheel dictates the robot’s heading and drives the robot. The front wheel is the focus of this research. This wheel will be referred to as the homeostasis enabling wheel (HEW), and the skin on the HEW will be referred to as the tegument throughout the remainder of the paper. The linkage that is rotated by Motor 3, and is attached to the drive shaft and the wheel will be referred to as the intermediate linkage for the remainder of the paper. Previous experimental research \cite{epps_feron_2019} on the HEW demonstrated that the wheel successfully rotates with a tegument attached to the system. For this paper, the HEW is analyzed without the tegument attached because the mechanics and feasibility of using the HEW as a form of locomotion are the main focuses of this paper.  
\begin{figure}[h!]
\center
    \minipage{0.2\textwidth}
        \includegraphics[width=\linewidth]{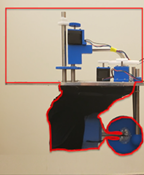}
    \caption{Outline of how the system would be enclosed}
    \label{OL}
    \endminipage\hfill
    \minipage{0.28\textwidth}
        \includegraphics[width=\linewidth]{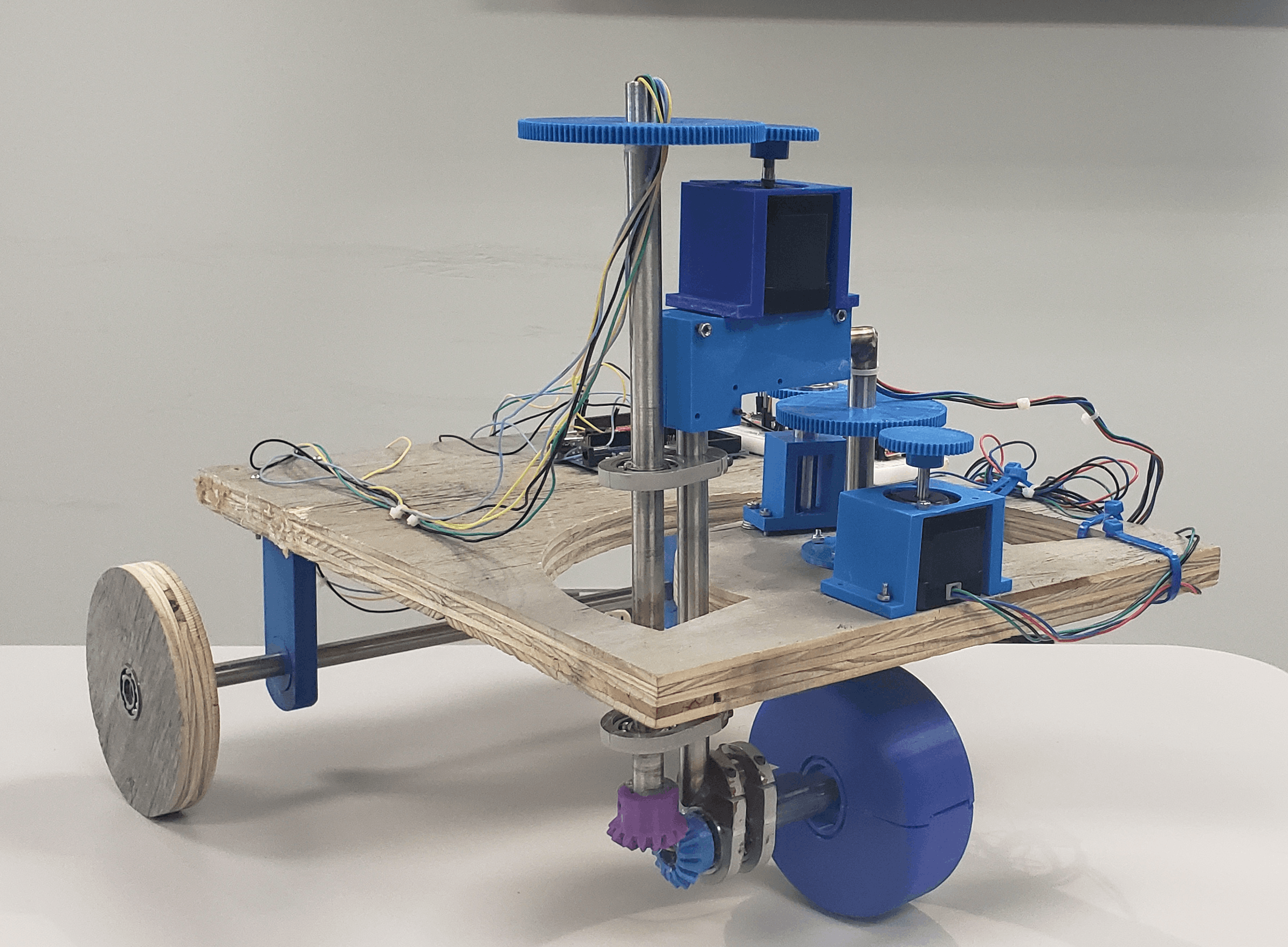}
        \label{SP}
        \caption{Starting orientation of the HEW on the robot}
    \endminipage\hfill
\end{figure}
Fig.~\ref{OL} shows where the tegument would be placed for a system using the HEW.

There are three motors on the robot that create a continuous forward rotation of the wheel. The first motor, Motor 1, provides torque for the wheel to rotate while Motor 3 rotates the linkage that is attached to the drive shaft and the wheel between $+90^{o}$  and $-90^{o}$ degrees.  Motor 1 oscillates the driveshaft between $+360^{o}$ and $-360^{o}$. The driveshaft is rigidly attached to an aluminum tube that is held by two ball bearings inside the wheel, requiring the aluminum tube to rotate with the driveshaft and the intermediate linkage. Inside the wheel, attached to the aluminum tube, is a servo motor, Motor 2, that controls the wheels angle relative to the driveshaft.  Motor 2 and Motor 3 must rotate simultaneously so that the wheel’s heading does not change as the intermediate linkage rotates about the wheel. For the robot to move along a straight path, the oscillatory motions of all three motors must be synchronized so that the wheel and the intermediate linkage are perpendicular to one another before Motor 1 drives the wheel. A video depicting the motion of the wheel can be found in \cite{epps_feron_2019}. 
\begin{figure}[h!]
\includegraphics[width=8cm]{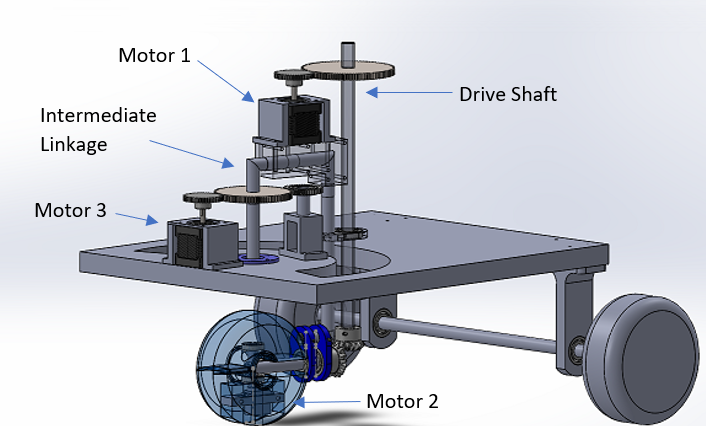}
\label{Labeld_Mod}
\caption{Computer Aided Design(CAD) model of proposed robot}
\end{figure}

The motion required to rotate the wheel twice while maintaining the integrity of the tegument is described in {\bf Algorithm ~\ref{algo}}. During the execution of {\bf Algorithm ~\ref{algo}} the tegument will be twisted about the driveshaft as the wheel rotates forward. The next motion involves the axle rotating to the opposite side of the wheel followed by the wheel rotating forward once again, which unbinds the tegument. These movements are executed consecutively.  In previous research\cite{epps_feron_2019} it was shown that a lightweight, flexible tegument with relatively low elastic resistance worked best regarding the wheel’s overall mobility. 

\begin{algorithm}[h!]
\caption{Function for two forward wheel rotations of the HEW.}\label{algo}\footnotesize
{\bf Function} rotate\_wheel\_twice\;
{\color{red} \% Assume the HEW is in the configuration of\\\% Fig.~\ref{SP}}\\
Rotate Motor 1 so that the driveshaft rotates +360 degrees\;
{\color{red}  \% The wheel turns a full +360 degree forward \\}
Rotate the frame and Motor 3 simultaneously  180 degrees\;
{\color{red} \%The intermediate linkage is now on the opposite side of the wheel, \\\% the wheel is still facing forward \\ }
Rotate Motor 1 so that the driveshaft rotates -360 degrees;\\
{\color{red} \%The wheel rotates forward;\\}
Rotate the frame and Motor 3 simultaneously -180 degrees; \\
{\color{red}The HEW is in the configuration of Fig.~\ref{SP} }\\
{\bf return}\;
\end{algorithm}

\subsection{Mathematical Consideration}

The principle known as Dirac\textquotesingle s Belt Trick\cite{1001.1778} or Plate Trick describes why the tegument remains fully intact as the wheel completes two forward rotations. Dirac’s belt trick simply illustrates that a 360 degree rotation is not topologically equal to no rotation, while a 720 degree rotation is. The proposed wheel has parallels with rotations SO(3), which has the known issue of “Gimbal Lock”. The relation between "Gimbal Lock" and the presented robot will be explored in Section VII of the paper.

\section{Characterization of the Motion of the Robot}
\subsection{Robot Position}

To discuss the kinematics of the robot, the general coordinates, as well as other notation will be defined in this section. Given a frame fixed to the surface being driven, $\mathcal{F}_{I} \in \mathbb{R}^{3}$,  $\{o_{I},\hat{i}_{I},\hat{j}_{I}, \hat{k}_{I}\}$, a reference point $P$ can be defined. Point $P$ is located between the two rear wheels of the robot. The origin of the robot's body-fixed frame, $\mathcal{F}_{rb} \in \mathbb{R}^{3}$, $\{o_{rb},\hat{i}_{rb},\hat{j}_{rb},\hat{k}_{jb}\}$ is located at reference point $P$. The pose of the robot can be described using the following variables:
\begin{itemize}
\item $x,y$ : The coordinates of reference point $o_{rb}$ in $\mathcal{F}_{I}$
\item $\rho$ : The orientation of the basis $\{\hat{i}_{rb},\hat{j}_{rb}\}$ with respect to $\mathcal{F}_{I}$
\end{itemize}
\begin{figure}[h!]
    \centering
    \includegraphics[width=5cm]{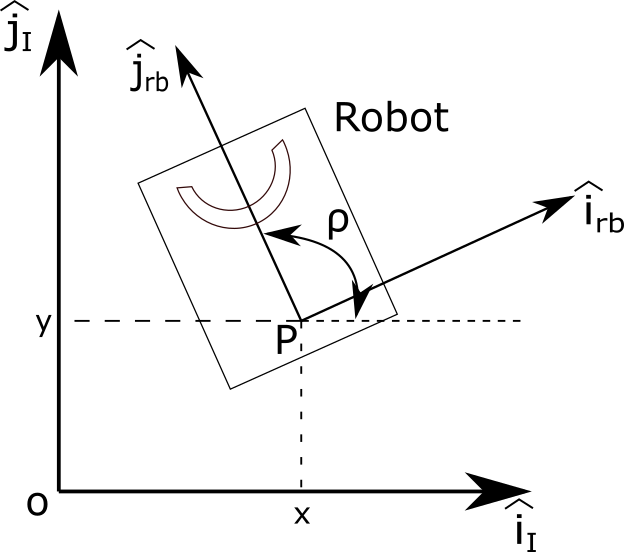}
    \caption{The robot's position relative to inertial frame}
    \label{fig:my_label}
\end{figure}

\subsection{Characterization of the Motion of the HEW}

\begin{figure}[h!]
    \centering
    \includegraphics[width=8cm]{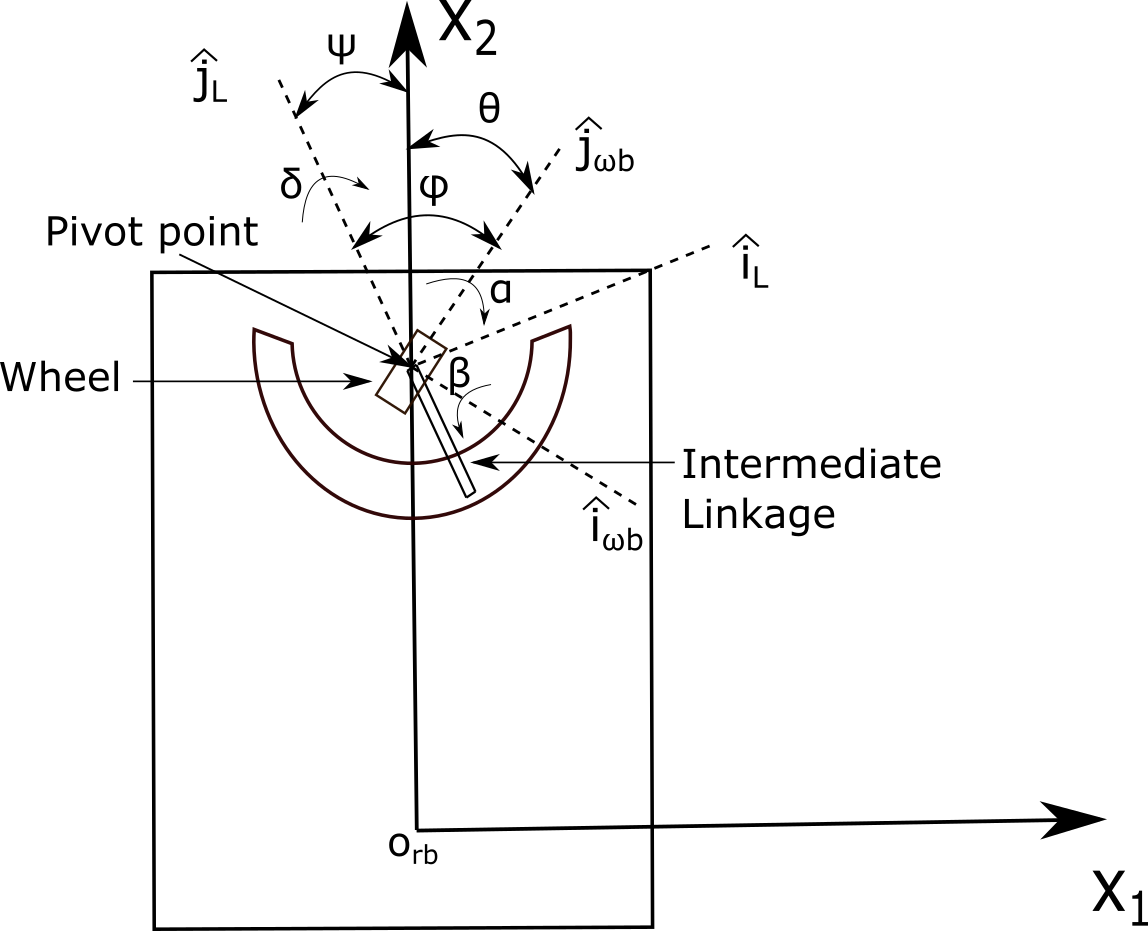}
    \caption{The angles that describe the motion of the HEW}
    \label{PPT}
\end{figure}

The rotation of the wheel and intermediate linkage can be characterized, as shown in Fig.~\ref{PPT}. The pivot point of the intermediate linkage is coincident with the $\hat{i}_{rb}$ axis of the robot’s body-fixed frame. Consider a reference frame, $\mathcal{F}_{l} \in \mathbb{R}^{3}$, $\{o_{l},\hat{i}_{l},\hat{j}_{l},\hat{k}_{l}\}$, that rotates with the intermediate linkage and has its origin fixed at the pivot point of the "Frame" . Underneath the chassis of the robot, the pivot point of the wheel is coincident with the $\hat{k}_{l}$ axis. Consider two reference frames, the first, $\mathcal{F}_{\omega b} \in \mathbb{R}^{3}$, $\{o_{\omega b},\hat{i}_{\omega b},\hat{j}_{\omega b},\hat{k}_{\omega b}\}$, with it's origin at the center of the wheel, that rotates with the wheel and another, $\mathcal{F}_{\omega f} \in \mathcal{R}^{3}$ with the basis $\{o_{\omega f},\hat{i}_{\omega f},\hat{j}_{\omega f},\hat{k}_{\omega f}\}$, with its origin fixed to the center of the wheel, and does not rotate with the wheel. The angle between $\hat{j}_{l}$ axis of the intermediate linkage's body-fixed frame and the $\hat{j}_{rb}$ axis of the robots body-fixed frame is denoted $\psi$, while the angle between $\hat{j}_{\omega b}$ and the $\hat{j}_{rb}$ axis of the robot’s body-fixed frame is denoted $\theta$.  The relative angle between the $\hat{j}_{l}$ axis and the $\hat{j}_{\omega b}$ axis can be described using the variable $\phi$, thus $\phi=\psi-\theta$.  The $\hat{j}_{l}$ axis is coincident with the driveshaft; therefore, the rotation about this axis is denoted $\delta$. 
 The angle between the $\hat{j}_{\omega b}$ axis and the $\hat{j}_{\omega f}$ axis created by rotating about the $\hat{i}_{\omega b}$ axis is denoted $\beta_{1}$, while the rotation about the axis $\hat{j}_{\omega b}$, the angle between the $\hat{i}_{\omega b}$ axis and the $\hat{i}_{\omega f}$ axis, is labeled $\alpha$. To fully characterize the motion of the robot, the angle of rotation of the two rear would need to be denoted. Due to the focus of this paper being the HEW, the angles of the two rear wheels will be neglected. The following vector of 8 generalized coordinates allow for a complete description of the robot's motion:
\begin{equation}
    q(t)=(x\: y\: \rho\: \theta\: \alpha\: \beta_{1}\: \psi\: \delta)
    \label{GenCord}
\end{equation}
Given the generalized coordinates, the configuration space of the robot is $SE(2)$ x $SO(3)$ x $S^{1}$.

\section{Kinematics}

\begin{figure}[h!]
    \centering
    \includegraphics[width=3cm]{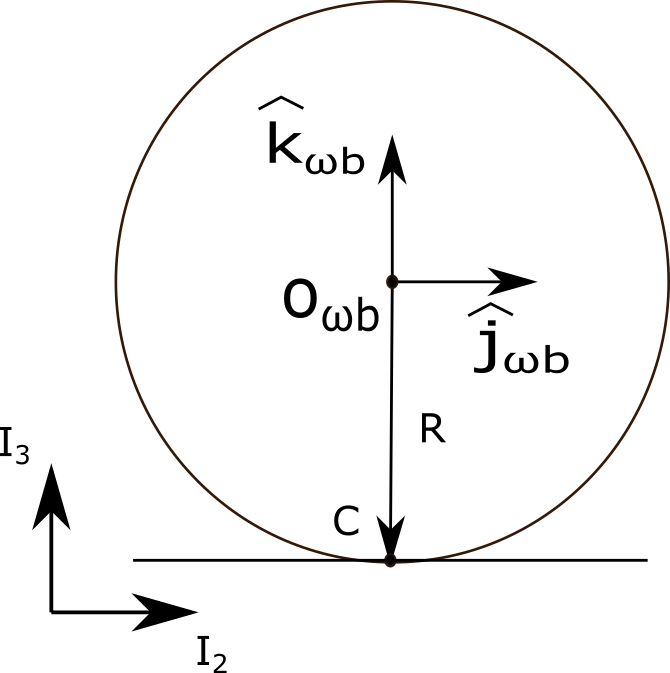}
    \caption{View of the wheel from the $\hat{i_{L}}\hat{k_{L}}$ plane given $\phi=90$, $\rho=90$, and $\theta=0$}
    \label{Wheel_diagram}
\end{figure}

Using the vector of generalized coordinates, Eq.~\ref{GenCord} , the kinematic equations describing the robot and the HEW can be written as Eq.~\ref{kin}.
\begin{gather}
    \begin{bmatrix} \dot x \\ \dot y \\ \dot \rho \\ \dot \theta \\ \dot \beta \\ \dot \alpha \\ \dot \psi \\ \dot \delta \end{bmatrix}=\begin{bmatrix} R\sin(\zeta)\cos(\rho) & 0 & 0\\ R\sin(\zeta)\sin(\rho) & 0 & 0\\ -\frac{R\sin(\zeta)}{d} & 0 & 0\\0 & 1 & -1 \\ \sin(\phi) & 0 & 0 \\ \cos(\phi) & 0 & 0\\ 0 & 0 & 1 \\ 1 & 0 & 0
    \end{bmatrix}
    \begin{bmatrix} u_{1} \\ u_{2} \\ u_{3}\end{bmatrix}
    \label{kin}
\end{gather}
The inputs, $[u_{1} \: u_{2} \:  u_{3}]^{T}$, correspond to the angular velocities of the drive shaft ($\dot \delta$), Motor 2 and the frame about its pivot point respectively. The variable $R$ represents the radius of the HEW wheel, and the variable $d$ represents the distance between point $P$ and the pivot point of the wheel. These kinematics assume that the wheel is infinitely thin. Without a proper control system, disturbances from gear slip, friction between the wheel and the ground, or noise from mechanical failures can cause $\phi \neq \pm90^{o}$. The motion created when $\phi \neq \pm90^{o}$, and Motor 3 drives the wheel, resembles a wobble. To properly describe how the kinematics of the robot are effected by the wobble motion of the wheel, the variable $\zeta$ is defined as follows in Eq.~\ref{piecewise} and Eq.~\ref{zeta}.
\begin{equation}
    \zeta(t)=
    \begin{cases}
        \phi(t)=\pm90, \hfill \zeta(t)=\psi(t)\\
        \phi(t)\neq \pm90, \hfill Eq.\ref{zeta}
    \end{cases}
    \label{piecewise}
\end{equation}
\begin{equation}
    \zeta(t)=cos^{-1}\bigg(\frac{\vec{v}\cdot \vec{w}}{\norm{\vec{v}}\cdot \norm{\vec{w}}}\bigg)sgn(\phi)
    \label{zeta}
\end{equation}
Consider Fig.~\ref{Wheel_diagram}. The point at which the wheel contacts the ground at a given $\delta(t)$  is labeled $C(t)\in\mathbb{R}^{3} $. Point $C(t)$ can be represented in the body frame of the wheel as $C(t)^{\omega f}=\begin{bmatrix}c_{1},\hfill  c_{2}, \hfill c_{3} \end{bmatrix}^{T}$. Viewing the wheel from the $\hat{i_{L}}\hat{k_{L}}$ plane when $\phi \neq \pm 90$, the wheel resembles an ellipse.
\begin{equation}
    \frac{(xcos(\delta(t))-y(x)sin(\delta(t)))^{2}}{a^{2}}+\frac{(xsin(\delta(t))+y(x)cos(\delta(t)))^{2}}{b^{2}}=1
    \label{ellipse}
\end{equation}
Implicitly deriving the general equation of an ellipse, Eq.~\ref{ellipse}, for the $x$ coordinate point results in an equation that, along with Eq.~\ref{ellipse}, can be used to find the $c_{2}$ and $c_{3}$ coordinates, with respect to the fixed frame of the wheel, of point $C(t)$. Basic trigonometry can then be used to find the $c_{1}$ coordinate of point $C(t)$. Therefore point $C(t)$ can be found given any $\delta(t)$ and $\theta(t)$, which then allows for the trajectory of $C$ to be found.

\begin{equation*}
    a=R\cos(\phi(t)) \hspace{5mm} b=R
    \label{ab}
\end{equation*}

 Using Eq.~\ref{g1}-~\ref{g2}, the path of point $C$ is a concatenation of vectors, $\vec{g}(t)=[g_{1}\hfill g_{2} \hfill g_{3}]^{T}$, of which orthogonal vectors $\vec{q}(t)$ can be found, Eq.~\ref{dot}.  

\begin{equation}
    g_{1}=\begin{cases}
        c_{1}(t)-c_{1}(t-1), \hfill \theta<0\\
        -(c_{1}(t)-c_{1}(t-1)),\hfill \theta>0
        \label{g1}
    \end{cases}
\end{equation}
\begin{equation}
    g_{2}=\sqrt(s^{2}-g_{1}^{2}),\hfill s=\frac{ \delta(t)-\delta(t-1)}{360}2\pi R
    \label{g2}
\end{equation}

\begin{equation}
    \vec{q}\cdot \vec{g}=0
    \label{dot}
\end{equation}
\begin{equation}
    \vec{v}=R_{z}\big(-90+\rho(t)+(-90+|\psi(t)|)sgn(\phi(t))\big)\vec{q}
    \label{v}
\end{equation}
\begin{equation}
    \vec{w}=R_{z}(-90+\rho(t))\begin{bmatrix} 0 \\ 1 \\ 0
    \end{bmatrix}
    \label{w}
\end{equation}
\begin{equation}
    R_z{\theta}=
   \begin{bmatrix} \cos(\theta) & -\sin(\theta) & 0\\ \sin(\theta) & \cos(\theta) & 0\\0 & 0 & 1
    \end{bmatrix}
\end{equation}
The unit vector $\vec{w}$ is the $\hat{j_{rb}}$ basis of the robot's body carried frame with respect to an inertial frame, which can be found using Eq.~\ref{w}. Using Eq.~\ref{v}, the vector $\vec{q}$ can be represented in $\mathcal{F_{I}}$ as $\vec{v}$.
The angle between vector $\vec{v}$, and $\hat{j_{rb}}$, denoted $\vec{w}$, is the value of $\zeta(t)$.

\section{Experimental Validation}

To validate the feasibility of the HEW as a mode of transportation, the tricycle seen in Fig.\ref{Labeld_Mod} was fabricated and tested. The test called for the tricycle to follow the logic seen in {\bf Algorithm 1}, causing two complete rotations of the wheel. Two Aruco markers\cite{Aruco2014} were placed on the vehicle, one of the drive shaft and the other on the chassis of the robot between the rear two wheels. These markers allowed the movement of the robot to tracked as well as the movement of the driveshaft.  An Inertial Measurement Unit (IMU) was placed inside the wheel, which captures the roll, pitch, and yaw of the wheel relative to an inertial reference frame. 
\begin{figure}[h!]
    \centering
    \includegraphics[width=6cm]{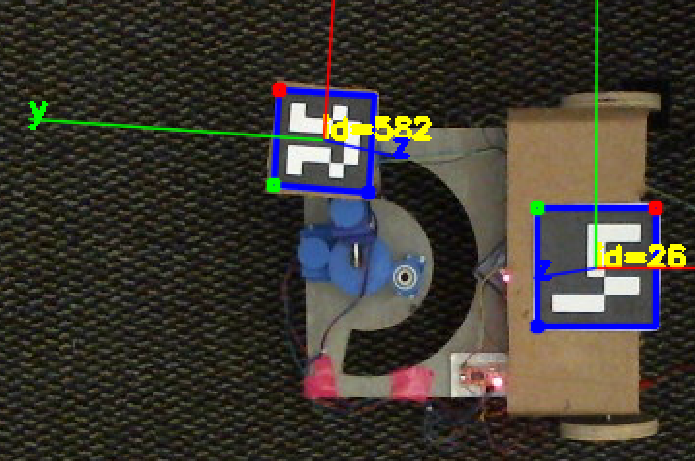}
    \caption{Aruco Marker Placement}
    \label{fig:my_label}
\end{figure}

\begin{figure}[h!]
\center
   \hfill \minipage{0.26\textwidth}
        \includegraphics[width=\linewidth]{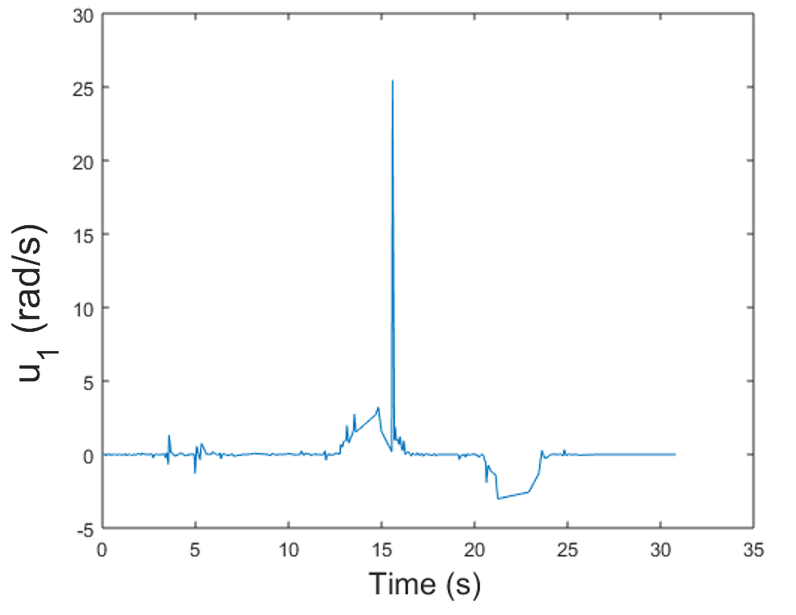}
    \caption{$\dot \delta$ vs. time}
    \label{w1}
    \endminipage
    \minipage{0.26\textwidth}
        \includegraphics[width=\linewidth]{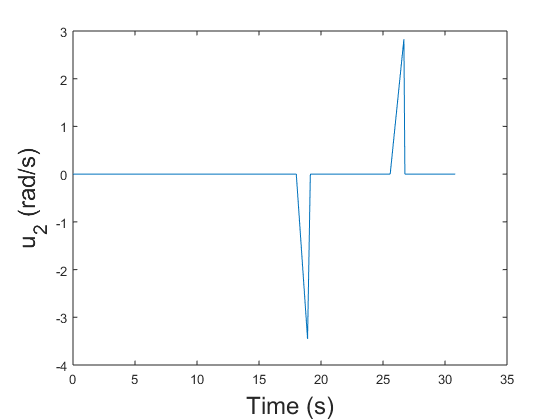}
        \caption{$\dot \theta$ vs. time} 
        \label{w2}
    \endminipage\hfill
    \minipage{0.27\textwidth}
        \includegraphics[width=\linewidth]{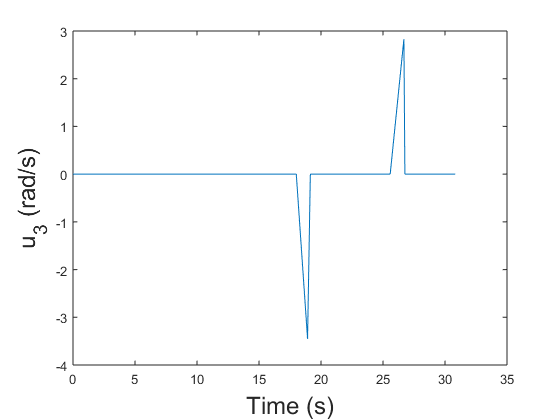}
        \caption{$\dot \psi$ vs. time}
        \label{w3}
    \endminipage
\end{figure}

\begin{figure}[h!]
\center
    \minipage{0.34\textwidth}
        \includegraphics[width=\linewidth]{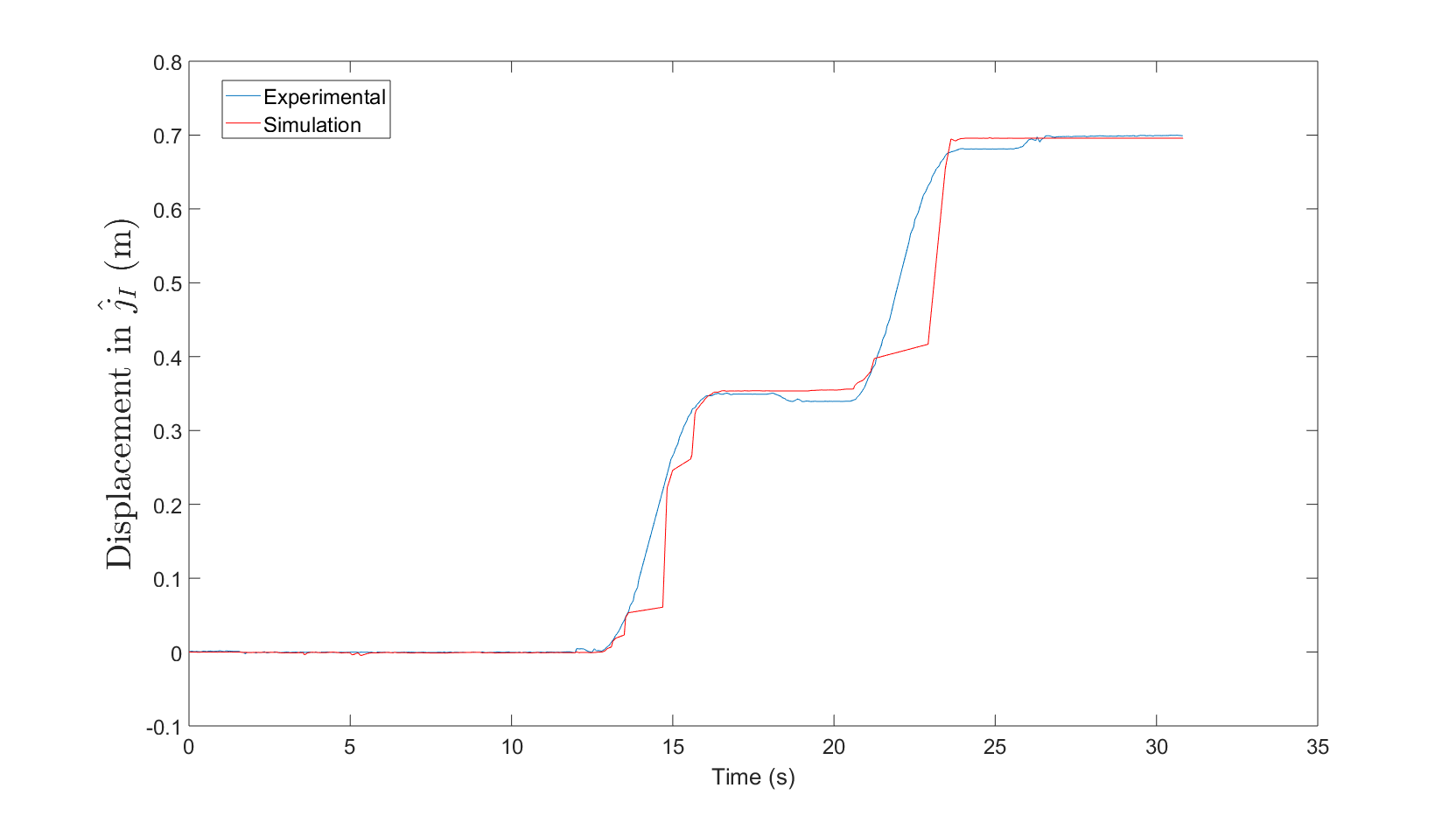}
    \caption{Displacement in $\hat{j}$ vs. time}
    \label{cy}
    \endminipage\
    \minipage{0.28\textwidth}
        \includegraphics[width=\linewidth]{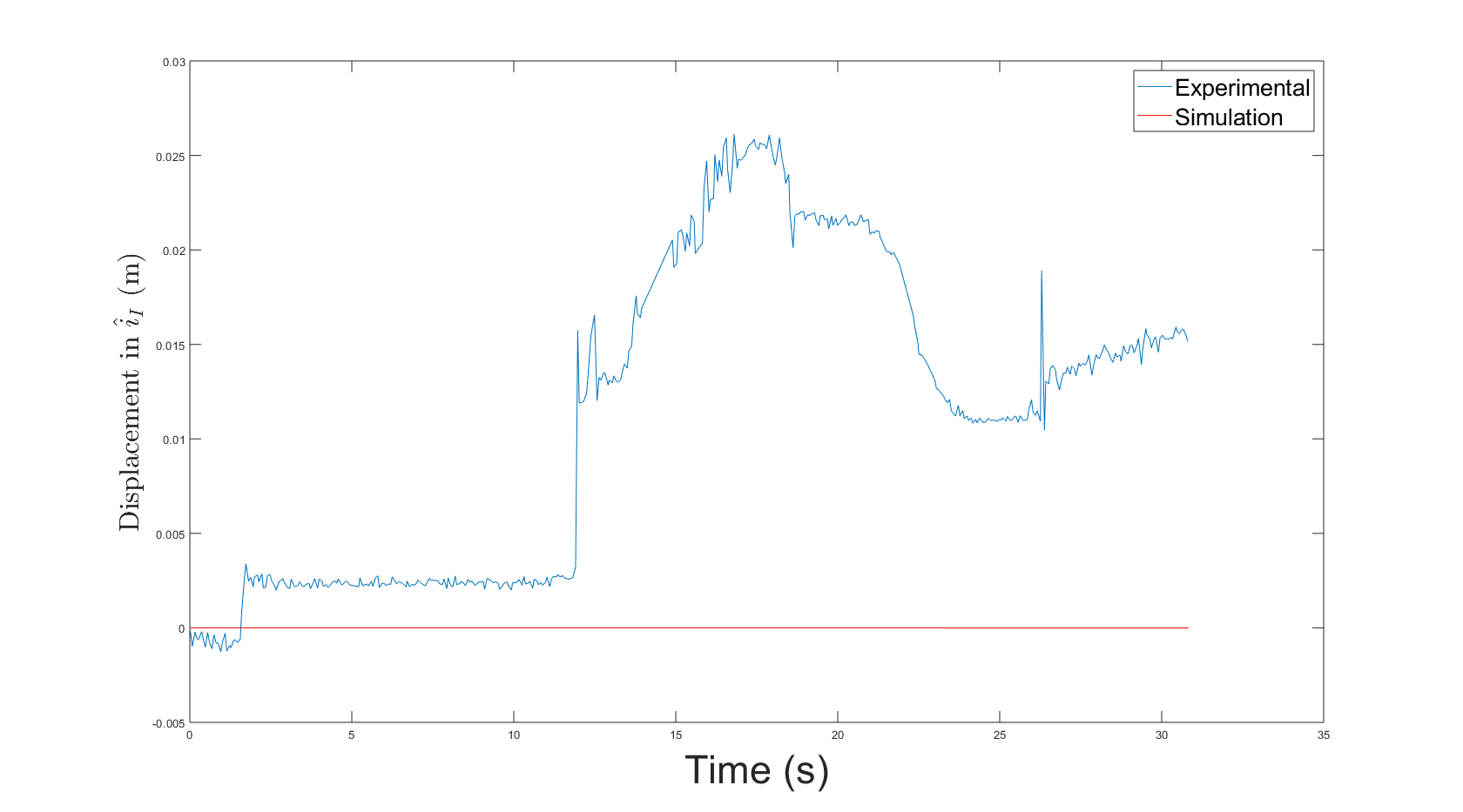}
        \caption{Displacement in $\hat{i}$ vs. time} 
        \label{cx}
    \endminipage
    \end{figure}

\begin{figure}[h!]
    \minipage{0.28\textwidth}
        \includegraphics[width=\linewidth]{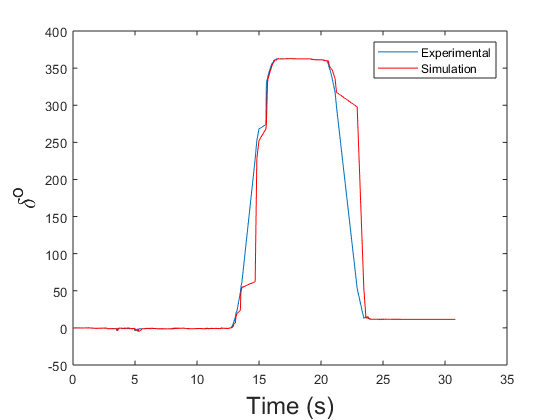}
        \caption{$\delta$ vs. time}
        \label{cd}
    \endminipage
    \minipage{0.25\textwidth}
        \includegraphics[width=\linewidth]{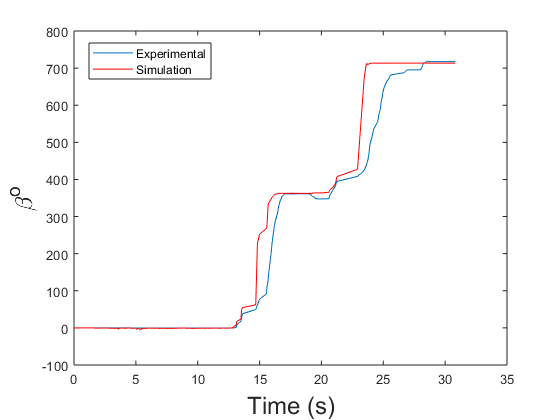}
        \caption{ $\beta$ vs. time}
        \label{cb}
    \endminipage
\end{figure}

The angular velocity and motion of the driveshaft and chassis were captured during the experiment and used as inputs into the simulated model created using Eq.\ref{kin}. These inputs can be seen as a function of time in Figs. \ref{w1}-\ref{w3}. 
The comparison between the expected motion of the robot and the actual motion of the robot can be seen in Fig. \ref{cy}-\ref{cb} thus validating the kinematics. The deviation seen in Fig.~\ref{cy}-~\ref{cb} is mainly due to noise caused by the friction between the wheel and the surface.

\section{Sensitivity Analysis}

\begin{figure}[h!]
    \centering
    \includegraphics[width=8cm]{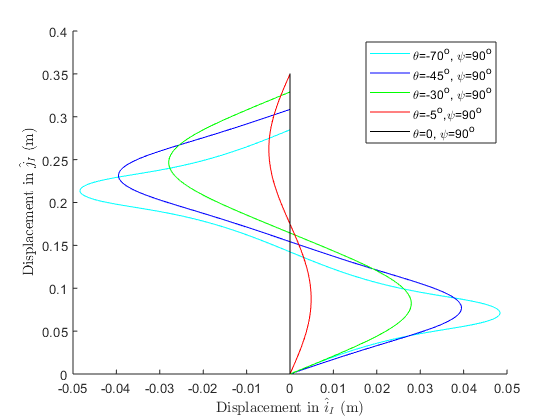}
    \caption{The path of point $C$ given $\rho=90$}
    \label{deviation}
\end{figure}

It can be observed that the direction that the robot will move in is orthogonal to the drive shaft when the value of $\phi=\pm90$. As previously stated, when $\phi \neq \pm90$ the path of the contact point between the infinitely thin wheel and the ground will resemble a wheel that wobbles, thus deviating away from the direction orthogonal to the drive shaft. This deviation is a function of $\phi$, while the orientation of the path in the inertial frame depends on $\rho$ and $\psi$. Fig.~\ref{deviation} illustrates the path of point $C$ in the case where $\theta \neq 0$.  
In the case where $\psi \neq \pm90$ similar deviations will be observed but the orientation of the path will be rotated as seen in Eq.~\ref{v}.

\section{Geometric Constraints and Limitations}

Although this design of a HEW fosters homeostasis, the design creates limitations that traditional wheeled vehicles do not have. To keep the tegument intact, the wheel cannot continuously rotate forward, the rotation of the drive shaft must be constrained. Therefore, the vehicle cannot reach its destination within a duration comparable to a traditional wheeled robot. Unlike a traditional wheel and axle with a limit on its angular velocity being dictated by the motor\textquotesingle s output, the flexibility and elastic resistance of the tegument can cause limit the angular velocity of the wheel well before reaching the motor\textquotesingle s maximum output. An obvious limitation of this design is that the angle φ must be $+90^{o}$ or $-90^{o}$ before torque is applied to the wheel for the contact point between the wheel and the ground to follow a straight-line path. From multiple experiments, it has been shown that when $\phi \neq \pm90^{o}$ and torque is applied to the wheel, the weight of the vehicle directed to the center of the wheel may cause the wheel to tip due to the inability of Motor 2 to resist the torque created by the weight of the robot. Due to the rotational inertia of the "Frame" being directly related to the mass of Motor 1, the mass of Motor 1 is coupled with the necessary torque for Motor 3 to successfully turn the frame without overshooting the desired angle $\psi$. Through observation, it can be seen that $\psi$ is geometrically constrained to be within the range of $+100^{o}$ and $-100^{o}$.  Although $\psi$ is constrained to $+100^{o}$ and $-100^{o}$, $\theta$ has a wider range of motion from $+180^{o}$ to $-180^{o}$. As mentioned in Section IIA, the HEW can experience "Gimbal Lock" if $\phi=0$. In this state, the similarity between "Gimbal Lock" in a gyroscope and "Gimbal Lock" in the HEW can be recognized due to the rotation axis of the drive shaft $f_{2}$ is aligned with the $w_{b2}$ axis of the body fixed frame of the wheel. Though this design presents limitations not experienced by a traditional wheel, there are several advantages of using the HEW.

\section{Applications and Advantages}

The presented system is only the architecture and mechanics needed for a wheeled robot to maintain homeostasis throughout the entirety of the system.  For the robot to enact and maintain homeostasis the addition of several sensors, actuators, a homeostatic control system, tasks, homeostatic responses, and homeostatic variables would be necessary.  Ownens\textquotesingle \cite{10.1007/978-3-540-73922-7_19} more thoroughly discusses what’s needed to enact homeostasis.  The presented robotic architecture would aid an artificial immune system in enabling homeostasis.  The HEW allows any part of the robot to be accessed internally through a network of tubes, hoses and wiring, which in turn enables the robot to repair virtually any part of itself unlike traditional robots with rotary elements.  This attribute of self-healing, can be advantageous for robots placed in areas that are not accessible to humans, such as the Mars Rover, Curiosity.  Other applications of the HEW include robots used to clean areas doused with high levels of radiation such as the nuclear reactors in Chernobyl\cite{abouaf_1998} and Fukashima.  Though the focus of this research was the mechanics of a homeostasis enabling wheel, the principle of creating the continuous rotation of an element while maintaining a fully connected interior can be applied to rotary elements of aquatic robots.

\section{Evolutionary Consideration}

The design of the proposed wheel adheres to the constrains of natural evolution and to the constraints\cite{biopolyverse_2015} necessary for biological systems to create and maintain homeostasis. By abiding by these constraints, one could infer that wheeled limbs could have become a mode of transportation developed through natural evolution. However, the evolutionary path to wheeled limbs would have had to have several iterations of "usable" and "desirable" wheel designs that lead to the wheel design presented in this paper. 

To argue against the claim that the HEW could not be created by the natural evolution of organisms on Earth, a possible biological implementation of the HEW presented in this paper will be discussed. The description of the biological implementation of the presented wheel may contain artificial equivalents to natural components that require more research to better mimic the system presented in this paper.

The components of a biological design are comparable to the aforementioned design illustrated in Fig. 2. These components consist of muscles, tendons, a skeleton and, a flexible membrane.  The biological implementation would require joints such as a shoulder, elbow, and wrist where the mechanical design has servos and pivot points. The above-mentioned biological implementation is only a preliminary consideration.
\begin{figure}[h]
\begin{center}
	\includegraphics[width=8cm]{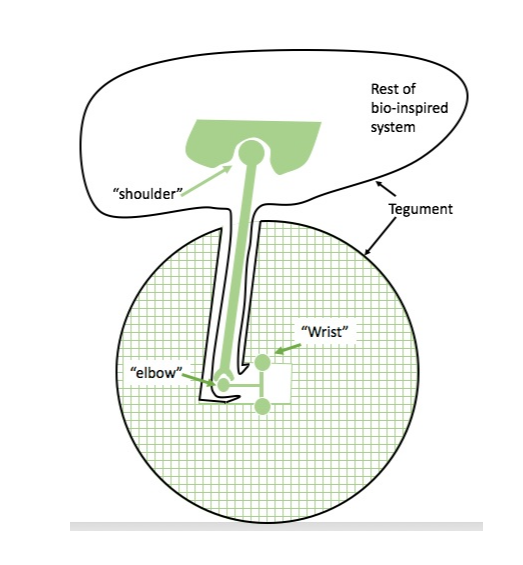}
\caption{Biological homeostasis-enabling wheel}
\label{Bio}
\end{center}
\end{figure}

The same constraints that must be placed on the mechanical HEW must also be placed on the biological design so that the wheel follows a straight line path. The implementation of a biological homeostasis-enabling wheel will rely heavily on the ability to fabricate muscle tissue and tendons that can perform these tasks.   

\section{Conclusion}

This paper presented a novel non-holonomic wheeled robot that has the necessary mechanics for a continuous tegument to enclose all of the internal parts of the robot, thus creating a fully connected interior throughout the entirety of the robot.  The results of this research provide a foundation for the idea that, given the proper environmental influences, evolution could have created higher-order life forms with wheeled limbs that are similarly designed to the wheel and axle that humankind has developed. The mechanics of the presented robot have been quantified, and the oscillatory motions of its motors that create a continuous forward rotation of the wheel were described. The next steps of this research consist of fitting the robot with a tegument, which will create the sub-region necessary for homeostasis. Lastly, homeostasis will be demonstrated by experimentally demonstrating that internal parameters of the robot can be regulated despite the presence of an external disturbance.  Other improvements on the design of the HEW will be implemented to decrease the time needed for the frame to rotate about the wheel. A controller along with encoders on the motors can be added to the HEW, which will reduce the error caused by noise when rotating the three motors.  This design of a HEW can be used as a proof of concept, creating the foundation for bio-inspired robotics that incorporates self-regulating rotating limbs.



%
\bibliography{Sections/bibliography}
\bibliographystyle{ieeetr}

\end{document}